\documentclass[conference,compsoc]{IEEEtran}
\usepackage{algorithm}
\usepackage{algorithmic}
\usepackage{array}
\usepackage{url}
\usepackage{xfrac}
\usepackage{float}
\usepackage{flushend}
\usepackage{enumerate}
\usepackage{dsfont}
\usepackage{nicefrac}
\usepackage{amsmath}
\usepackage{bm}
\usepackage{graphicx}

\newcolumntype{M}[1]{>{\centering\arraybackslash}m{#1}}
\newcommand{\argmin}{\mathrm{arg}\displaystyle\min}

\newcolumntype{N}{@{}m{0pt}@{}}

\ifCLASSOPTIONcompsoc
\usepackage{cite}
\else
  \usepackage{cite}
\fi

\begin{document}

\title{Lossless (and Lossy) Compression of \\Random Forests}


\author{\IEEEauthorblockN{Amichai Painsky}
\IEEEauthorblockA{School of Computer Science and Engineering\\
The Hebrew University of Jerusalem\\
Givat Ram, Jerusalem 91904, Israel\\
Email: amichai.painsky@mail.huji.ac.il}
\and
\IEEEauthorblockN{Saharon Rosset}
\IEEEauthorblockA{Department of Statistics\\
Tel Aviv University\\Tel Aviv 69978, Israel\\
Email: saharon@post.tau.ac.il}}

\maketitle

\begin{abstract}
Ensemble methods are among the state-of-the-art predictive modeling approaches. Applied to modern big data, these methods often require a large number of sub-learners, where the complexity  of each learner typically grows with the size of the dataset. This phenomenon results in an increasing demand for storage space, which may be very costly. This problem mostly manifests in a subscriber based environment, where a user-specific ensemble needs to be stored on a personal device with strict storage limitations (such as a cellular device). In this work we introduce a novel method for lossless compression of tree-based ensemble methods, focusing on random forests. Our suggested method is based on probabilistic modeling of the ensemble's trees, followed by model clustering via Bregman divergence. This allows us to find a minimal set of models that provides an accurate description of the trees, and at the same time is small enough to store and maintain. Our compression scheme demonstrates high compression rates on a variety of modern datasets. Importantly, our scheme enables predictions from the compressed format and a perfect reconstruction of the original ensemble. In addition, we introduce a theoretically sound lossy compression scheme, which allows us to control the trade-off between the distortion and the coding rate. 
\end{abstract}


%
\IEEEpeerreviewmaketitle

\section{Introduction}
An ensemble method is a collection of sub-learners, usually decision trees like CART \cite{breiman1984classification} or C4.5/C5.0 \cite{quinlan2014c4, quinlan2004data}. The ensemble takes advantage of the favorable properties of its sub-learners, while mitigating their low accuracy by averaging or adaptively adding together many trees. Widely used ensemble methods include bagging \cite{breiman1996bagging}, boosting \cite{schapire2003boosting}, random forests \cite{breiman2001random} and others. During the past decades ensemble methods have gained a wide reputation of being among the most powerful off-the-shelf predictive modeling tools \cite{hastie2009elements}. 

In order to attain their favorable predictive performance, ensemble methods usually require a large number of sub-learners, which tends to grow with the size of the problem. An increasing dataset size also results in deeper and more complex models. This most clearly manifests in random forest, where the trees are typically grown to a maximal size and are not pruned  \cite{breiman2001random}. Consequently, their size strongly depends on the number of observations. For example, training a random forest of $1000$ trees (using Matlab's {\sf treeBagger} routine) on a modern big dataset such as Liberty Mutual Group's Property Inspection Prediction\footnote{\url{https://www.kaggle.com}} (which consists of $50,999$ observations and $32$ features), results in an average tree depth of $40$ levels. Storing these trees require $733.7$ MB with the best standard solution (that is, using the {\sf compact(tree)} Matlab routine, followed by a gzip compression \cite{deutsch1996gzip}).  

In this work we present an extended version of \cite{painsky2016compressing}, which focuses on lossless compression method for large tree-based ensembles. The fundamental observation underlying our method is that the random forest's trees are independent and identically distributed random entities, given the training data. This allows us to infer their probabilistic structure and construct an entropy code with a corresponding dictionary. As later discussed, more complicated models better describe the true probabilistic structure of the trees and therefore result in better compression rates. However, such complicated models also result in codes which require a greater number of dictionaries (and henceforth increase the overall compressed data description). Therefore, the main challenge is finding the ideal tradeoff between an accurate description of the model and the total dictionary size.  Our compression approach is lossless in its essence. This means we allow complete recovery of the original trees without any loss of information. Moreover, with a careful implementation, our suggested approach allows prediction straight from the compressed format. 

In this extended version, we further introduce a novel lossy compression scheme which demonstrates a greater coding rate at the cost of a distortion in the reconstruction. Our lossy compression is based on subsampling and quantization of the ensemble trees, followed by lossless compression. This allows us to introduce a fundamental tradeoff between distortion and coding rate in i.i.d. ensemble methods. 

A matlab implementation of our suggested compression scheme is publicly available at the first author's web-page\footnote{\url{https://sites.google.com/site/amichaipainsky/software}}.

\subsection{Related work}
\label{related work}
The problem of storing large ensembles has gained an increasing interest in recent years, due to these methods' popularity and the emergence of extremely large datasets.

One line of work focuses on ``pruning" techniques for tree ensembles. Here, the idea is to reduce the size of the ensemble by removing redundant components (features/trees etc.), while maintaining the predictive performance.   
In \cite{geurts2000some}, the authors propose to extend the classical cost-complexity pruning of individual trees to ensembles. On the other hand,  \cite{meinshausen2010node,friedman2008predictive} propose to prune and improve the model's interpretability by selecting optimal rule subsets from tree-ensembles. Another way to reduce the complexity and/or improve the accuracy of the tree-ensembles is to merely select an optimal subset of trees from a very large ensemble generated in a random fashion (see, e.g. \cite{bernard2009selection}).
An additional pruning-based approach \cite{joly2012l1} is to reformulate the tree-ensemble as a linear model in terms of node indicator functions, while adding an $L_1$-norm regularization term (LASSO) to encourage sparsity in the features. The idea behind this approach is to select a minimal subset of indicator functions while maintaining predictive accuracy.  
Notice that all of these ``compression" schemes are lossy and result in a pruned ensemble which is significantly different from the original ensemble. Moreover, there are no guarantees on the combination of compression rate and the difference between the pruned and original ensemble. In other words, some ensemble may be successfully pruned while others may not.

In a different line of work Bucelia et al. \cite{bucilua2006model} suggest to ``compress" an ensemble model by training an artificial neural network that mimics the functionality of the ensemble. This results in a significantly faster and more compact approximation of the original model. Despite these favorable properties, approximating an ensemble by a neural network is again both lossy and irreversible. In other words, the neural network predictions are not identical to the predictions of the original ensemble and in some cases may deviate quite notably in terms of root mean square error. Moreover, given the approximated neural network, it is not possible to recover the original ensemble. This means that once an approximation network replaces the ensemble, one cannot make further use or modifications to the ensemble (for example, add more trees to improve performance). In addition, notice that for a modern big--data,  the trained random forest would usually consist of complex and deep (un-pruned) decision trees. It is well known that training a neural network to accurately approximate such a complicated function is not a trivial task. In fact, it typically requires an exponentially increasing number of neurons to achieve a prescribed accuracy \cite{tikk2003survey}.    

There also exists a large body of work on the compression of different data structures in the source coding community. This includes the compression of a single or multiple tree structures \cite{katajainen1990tree,chen1996cient}. However, this line of work focuses on more general settings - usually arbitrary or randomly constructed trees. These trees hold  different probabilistic characteristics than our data-driven decision trees, which are all built on a single dataset and with only the randomness infused by the random forest algorithm differentiating them. 

To the best of our knowledge, our contribution provides the first lossless compression approach for large tree ensembles. 

\section{Basics}
\subsection{Random forests}  
A random forest is an ensemble learning method, usually used for classification or regression problems \cite{breiman2001random}. It operates by constructing multiple decision trees at the training phase, followed by aggregating their results by a majority vote (classification) or averaging (regression). This overcomes the well-known drawback of a single decision tree, which tends to have low accuracy and high variance due to its greedy model building approach.
In a random forest, each tree is constructed according to a randomly sampled subset of observations (usually with replacement), and a randomly sampled set of variables. This allows a diverse set of learners  which is then averaged, thus reduces the variance associated with a single tree, and decreasing the generalization error.

Random forest's trees are usually constructed by widely-used tree fitting methods like CART \cite{breiman1984classification} or C4.5/C5.0 \cite{quinlan2014c4, quinlan2004data}. These methods are greedy recursive partitioning algorithms. In each iteration, a set of observations is split into disjoint subsets, such that a loss criterion \cite{painsky2018universality,painsky2018bregman} is minimized, in a greedy, non-regret manner (for example, \cite{breiman1984classification,hothorn2006unbiased, painsky2017cross}). A regression or classification tree is a tree data structure in which each internal (non-leaf) node is labeled with a variable name and a corresponding split value, while a leaf is labeled with a fitted value (a class for classification problems, or a numerical value for regression problems).

 \subsection{Entropy coding}
\label{entropy coding}
A compressed representation of a dataset involves two components -- the compressed data itself and an overhead redundancy. Encoding a sequence of a length $n$ requires at least $n$ times its empirical entropy. This is attained through entropy coding according to the sequence's empirical distribution.
The redundancy, on the other hand, may be quantified in several ways. 
One simple way is through a dictionary. Assume we encounter $n_0 \leq n$ unique symbols. Then a dictionary is simply a one-to-one mapping of each unique symbol and its corresponding codeword. 
An alternative way to quantify the redundancy is through a reference distribution. Assume we encode the source sequence according to a fixed (and predefined) distribution $Q$ while the empirical distribution is $P$. Then, the Kullback Leibler divergence of $Q$ from $P$, denoted $D_{KL}(P‖Q)=\sum P_i \log \frac{P_i}{Q_i}$, is the amount of information lost when $Q$ is used to approximate $P$. In other words, $nD_{kl}(P||Q)$ is the expected number of extra bits required to encode the $n$ samples from $P$ using a code optimized for $Q$ rather than the code optimized for $P$. Hence, the trade-off is between having efficient codes and large overhead (when using a detailed dictionary) or having inefficient codes with no overhead (when using a predefined reference distribution). 

There exist several popular entropy coding schemes. The most widely used are Huffman and arithmetic coding \cite{sayood2012introduction}.
The Huffman algorithm is an iterative construction of a variable-length code table for encoding the source symbols. The algorithm derives this table from the probability of occurrence of each source symbol. It can be shown that the average codeword length,  achieved by the Huffman algorithm, $R$, satisfies $\hat{H}\left(X \right) \leq R \leq \hat{H}\left(X \right)+1.$ In arithmetic coding, instead of using a sequence of bits to represent each symbol, we represent it by a subinterval of the unit interval \cite{sayood2012introduction}. This means that the code for a sequence of samples is an interval whose length decreases as we add more samples to the sequence. Assuming that the empirical distribution of the sequence is known, the arithmetic coding  procedure achieves an average codeword length which is within $2$ bits of the empirical entropy. Although this is not necessarily optimal for any fixed sequence length (as the Huffman code), this procedure is incremental and can be used for any sequence-length.  One of the  major challenges of entropy coding occurs when the source is over a large alphabet size. Then, the coding redundancy becomes quite significant \cite{szpankowski2012minimax} and alternative compression methods should be considered \cite{orlitsky2004universal,painsky2015universal,painsky2016simple,painsky2017large,painsky2016generalized,painsky2018phd,painsky2018linear}.
 
In addition to the entropy coders discussed above, it is important to mention the Lempel-Ziv (LZ)-based family of coders \cite{sayood2012introduction}. LZ-based algorithms replace repeated occurrences of source sequences with references to a single copy of that sequence existing earlier in the uncompressed stream. The main advantage of this scheme is that it does not require to transmit a dictionary, nor a predefined reference distribution. Yet, the LZ-based algorithms' compression rate asymptotically approaches the empirical entropy of the sequence.

 \section{Compression methodology}
\label{compression methodology}
A tree-based ensemble is a collection of decision trees, usually like CART or C4.5/C.5.
Tree building algorithms can handle both numerical and categorical features and build models for regression, two-class classification and multi-class classification.
The splitting decisions in these algorithms are based on optimizing a splitting criterion over all possible splits on all variables. This means that each node in the constructed tree is defined by both a splitting variable and a corresponding split value. The fits of the tree are minimizers of the objective function for the resulting sets of leaf observations. For example, the fit of the observations in a certain leaf of a regression tree is simply the average value of these observations. 
A single tree structure may hold many additional characteristics and parameters (such as various summary statistics at each node). Since we are interested in compression for prediction purposes, we limit our attention to the following relevant attributes:    

\begin{enumerate}
\label{list}
\item The structure of the tree
\item The splits of the nodes (variable name and a corresponding selected split value)
\item The values of the leaves (fits)
\end{enumerate}
where the structure of the tree is simply a data-structure which distinguishes between nodes and leaves (for example, Figure \ref{fig:tree}).  
In this work we focus on the compression of random forests, in which the trees are constructed independently and are identically distributed, given the training data. In order to apply entropy based compression methods (such as Huffman or arithmetic coding) we first need to define a probabilistic setup for the entity we are to compress. We have that
\begin{align}
P(tree)=&P(tree\ structure)\cdot \\\nonumber
& P(nodes | tree\ structure) \cdot \\\nonumber
&P(leaves | nodes, tree\ structure).
\end{align}
This decomposition allows us to compress each of the components separately, while benefiting from a reduced algorithmic complexity.


\subsection{Tree structure compression}
\label{tree structure compression}
The problem of compressing a generalized tree-based data structure has received a considerable amount attention throughout the years \cite{katajainen1990tree}. 
Here we introduce an encoding method presented by Zaks \cite{zaks1980lexicographic}. However, there exist many other compact representation formats for the structure of a tree, as later described.

\begin{figure}[!ht]
\centering
\includegraphics[width = 0.62\textwidth,bb= 70 510 705 710,clip]{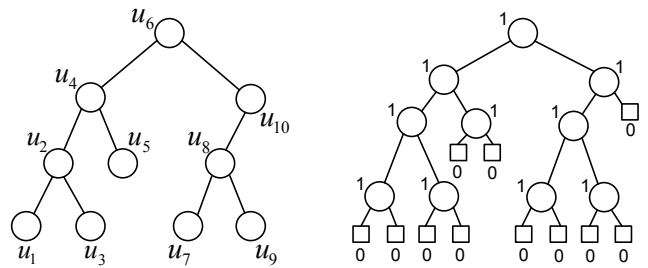}
\caption{Zaks' tree binary representation. Left: a decision tree. Right: The numbering of the nodes and the leaves related to Zaks' sequence}
\label{fig:tree} 
\end{figure}

Consider the tree in left chart of Figure \ref{fig:tree}. Label all the nodes by $1$ and all the leaves (missing subtrees) by $0$ as in right chart. We obtain the code sequence, called Zaks' sequence, by reading the labels in preorder (first visit the root, then recursively traverse the left subtree in preorder, and then the right subtree in preorder). Hence, the Zaks' sequence related to the tree in Figure \ref{fig:tree} is $111100100100111001000$. 

We have the following characterization for feasible Zaks' sequences. A bit string is a Zaks' sequence if and only if the following three conditions hold:
\begin{enumerate} [i]
\item The string begins with $1$
\item The number of $0$'s is one greater than the number of $1$'s
\item  No proper prefix of the string has the property ii.
\end{enumerate}
Hence, the length of a Zaks' sequence is $2n + 1$ for a tree with $n$ nodes and it is uniquely decodable \cite{zaks1980lexicographic}.

There exist several other tree structure encoding scheme \cite{katajainen1990tree}, such as \textit{children pattern sequence} (of length $2n$) and \textit{balanced parentheses} (again, of length $2n$) or others.

As shown in the following sections, the structure of the tree holds a relatively small size, compared with the other compressed components. Therefore, we choose to represent each of the trees' structure with a Zaks sequence, concatenate all sequences, and apply a simple LZ-based encoder \cite{sayood2012introduction} to the concatenated sequence. Notice we may have treated each Zaks' sequence as an independent realization from $P(tree\ structure)$ and encode accordingly. However, this approach would treat each sequence as a single symbol, drawn from a very large alphabet (of all possible sequences), and ignore the internal structure of the sequences. Therefore, inspired by \cite{chen1996cient}, we compress the concatenated sequence using an LZ-based encoder, and take advantage of the structural nature of Zaks' sequences.

\subsection{Nodes compression}
\label{nodes compression}
In this section we focus on the compression of the trees' nodes (specifically, the split selected at each node).
As mentioned above, each node is defined by a name of a variable and a corresponding split value.
Notice some variables may be numerical while others categorical, and the range of values of each variable may also be significantly different than the other variables. Therefore, we derive a probabilistic model for each of the variables separately. In addition, we notice that a node only depends on its parents, as a result of the recursive construction of the tree. This means that
\begin{align}
\label{nodes}
&P(nodes | tree\ structure) =\\\nonumber
&\prod_{u \in \{nodes\}} P(u \ variable \ name | u \ parents) \cdot \\\nonumber
&P(u \ split \  value | u \ parents, u \ variable \ name).
\end{align}

At this point it becomes quite evident that if we are to define a separate probabilistic model for each term in (\ref{nodes}), (for example, $P(u_2 \ \text{variable name} \  |\  \text{root name}, \ \text{root split value})$), we would end up with a number of models which is exponential in the depth of the tree. This phenomenon is further demonstrated in Section \ref{Clustering of node models}.
Moreover, encoding each node`s information according to its specific model would result in an exponentially increasing number of dictionaries, as discussed in Section \ref{entropy coding}. This means we need to ``cluster" models together, in order to reduce the dictionary size overhead, while maintaining a good compression rate.

\subsubsection{Model clustering} 
\label{model clustering}
Let $s_1, \dots s_M$ be $M$ sequences of independent draws, with corresponding empirical distributions $P_1,\dots,P_M$, all on the same alphabet. Denote the lengths of the sequences as $n_1,\dots, n_M$, respectively. We would like to encode all of these sequences according to a single codebook (and a single corresponding dictionary). 
Let $Q$ be the probability distribution according to which the codebook is constructed. Then, the minimal overhead redundancy, where the minimization is with respect to the probability distribution $Q$ is:
\begin{align}
\label{clust}
\min_Q \sum_{i=1}^M n_i D_{kl}\left(P_i||Q\right)+\alpha{||Q||}_0
\end{align}  
where $D_{kl}$ is the kullback leibler divergence (previously defined in \ref{entropy coding}), ${||Q||}_0$ is the $L_0$ norm of $Q$ (number of non-zero elements in $Q$) and $\alpha$ is the cost of describing a single line in the dictionary (a symbol and its codeword). The $L_0$ term makes this optimization problem quite involved. Therefore, we may relax it by replacing the $L_0$ term with  $L_1$ (Lasso-like) or $L_2$ (Ridge-like) penalties, to achieve a convex optimization problem.
Alternatively, assume that the alphabet size (from which each of the sequences is drawn) is finite and equals $B$. Then $||Q||_0 \leq B$ and the minimal value of (\ref{clust}) is bounded from above by
$\sum_{i=1}^M n_i D_{kl}\left(P_i||Q^*\right)+\alpha B$,
where $Q^*$ is the minimizer of $\sum_{i=1}^M n_k D_{kl}\left(P_i||Q\right)$.
Further, let us assume that the $B$ is fixed, while the lengths of the sequences ($n_1,\dots, n_M$) increase. In this case, the first term becomes dominant, compared to the penalty term,
$\alpha B \ll \sum_{i=1}^M n_i D_{kl}\left(P_i||Q^{*}\right)$.
This means that for a fixed $B$, and as the $n$'s increase, we may approximate the penalty term as a constant and replace (\ref{clust}) with 
\begin{align}
\min_Q\sum_{i=1}^M n_k D_{kl}\left(P_i||Q\right)+\alpha B
\end{align}

Let us now extend this problem and assume that the $M$ sequences are to be clustered according to $K$ different codebooks. For a fixed $K$, the corresponding optimization problem is
\begin{equation}
\label{clust2}
\min_{\underline{C},\underline{Q}} \sum_{k=1}^K \sum_{i=1}^M  \mathds{1}_{\{P_i \in C_k\}}n_i D_{kl}\left(P_i||Q_k\right)+\alpha{||Q_k||}_0
\end{equation}  
where $\underline{C}=\{C_i\}_{i=1}^K$ and $\underline{Q}=\{Q_i\}_{i=1}^K$ are the clusters and corresponding codebook probability distributions, and $\mathds{1}\{\cdot\}$ is the indicator function. As before, the penalty term may be bounded from above by $\alpha B$, which leads to 
\begin{align}
\label{clust3}
\min_{\underline{C},\underline{Q}} \sum_{k=1}^K \sum_{i=1}^M  \mathds{1}_{\{P_i \in C_k\}} n_i D_{kl}\left(P_i||Q_k\right)+\alpha B K.
\end{align}  
This means that for sufficiently large $n$'s and a fixed $B$, we may bound (\ref{clust2}) from above, to achieve a simple clustering problem (\ref{clust3}). Notice this clustering problem is very well studied \cite{banerjee2005clustering} with many algorithms (mostly K-means like) and applications.

\subsubsection{Clustering of node models}
\label{Clustering of node models}
As mentioned above, 
we would like to cluster models together, to find the ideal trade-off between a minimal number of dictionaries and a minimal loss of bits which results from encoding the models according to the cluster's codebook. As demonstrated in (\ref{nodes}), we distinguish between modeling the variables' names and modeling the split values, given the variable name.

Let us first focus on the modeling of variable names. We would like to assign a designated probability distribution for a variable name, for each node in the tree, and then cluster the distributions as in (\ref{clust3}).
We begin by defining an empirical distribution which describes the variable name in the root. Then, we may define an empirical distribution of the root's children given the root, and so forth.
Obviously, the number of distributions quickly becomes intractable as we go deeper in the tree, even before we apply the clustering. Therefore, we relax the exhaustive construction of all possible models and focus on a simpler form of dependencies in the tree, in which we assume a node only depends in its depth and the variable name of its father. Therefore, assuming a forest with a maximal tree depth $T$, the number of possible models for the variable name is $d \cdot T$. 

Once we have established the list of possible models for variable names, we are ready to cluster the models according to (\ref{clust3}), for different values of $K$, and choose the one which minimizes the objective. We then compress the data which corresponds to each model with a Huffman code, according to the cluster's empirical probability distribution.

Notice that the cost of describing a single dictionary line, defined as $\alpha$ in (\ref{clust3}), depends on the nature of the data we are to compress and the encoder we use. Here, we may achieve a reduced dictionary size by holding a single dictionary which maps the actual name of the variable to its numeric representation and use the numeric representation in all the dictionaries we construct (e.g. instead of using the variable names ``height", ``weight" and ``eye color" we use ``00", ``01", ``10").  Since we do not know the codeword used for each symbol in  the dictionaries, we may bound it by the maximal length of a codeword, which is $d$ bits (the worst--case Huffman codeword for an alphabet size $d$). Therefore, we have that $\alpha = \log_2(d)+d$ for the variable names.  

In the same manner we would like to model the split value, given the name of the variable. We use the same modeling relaxation and construct a model according to the same dependencies described above. This leads to a total of  $d^2 \cdot T$ candidate models for clustering, since we need a different model of split values for each of the variable names models defined above. Assuming that a variable's split values take over $C$ different values, then the maximal codeword length is $C$ bits and $\alpha = \log_2(C)+C$. Obviously, $C$ may may quite large for numerical variables. However, in most decision trees (such as CART or C4.5/C5.0), a numerical split is specified by a single observation's value. This means that the numerical split value may be represented by an index of an observation, which takes $\log_2(n)$ bits. This naive representation may be further improved by applying entropy coding to these split values, as previously demonstrated. Therefore, we have that for numerical split values, $\alpha = \log_2(n) + C$.       

At this point it is important to emphasize an additional difference between the split value of numerical and categorical variables. The split values of a numerical variable are numeric values. Therefore, the distribution of these values is continuous, and there is a natural order between every two different values. On the other hand, split values of a categorical variable are partitions of its categories into two disjoint sets. This means that there is no natural ordering and the distribution of the values is discrete (takes over a finite set). In other words, designing an entropy encoder (which is designated for a finite set of unordered symbols) is much more natural for categorical split values than numerical ones. However, notice that for large datasets, variables' split values tend to take over a limited set of values as we are closer to the root, for both numerical and categorical variables. This means we can regard the numerical values as categories in this sense. As we go deeper in the tree, the split values become more uniform (and sparse) for both categorical and numerical variables, so most coding techniques are ineffective. These phenomena are discussed in greater detail in the Experiments section.

\subsection{Fits compression}
\label{fits compression}
We now turn to consider the compression of the tree's fits. As in the nodes' compression, we may model the (conditional) probability of fit values in each leaf of the tree. However, this requires an exponentially increasing number of models, as demonstrated in the previous section. Therefore, we define a simplified model in which the distribution of the fits in a leaf depends on its depth and its father's variable name. This leads to a set of probability distributions which we cluster according to (\ref{clust3}), in the same manner mentioned above.
As before, it is important to distinguish between compressing numerical and categorical values. In a classification problem, the fits are categorical and take over a finite set of values. This makes the use of entropy coders very suitable. However, regression problems result in numerical fits which may take over a continuous (and ordered) set of values. This means that we may either ignore this property and treat the continuous fits as categorical, or quantize the fits (through simple rounding, or in a more more complicated manner using a frequency based quantization technique, such as Lloyd--max algorithm \cite{lloyd1982least}). 

Notice that by quantizing the fits we introduce an error from the original tree so we can no longer regard our method as lossless. However, such quantization results in a very regularized distortion, in the sense that we can directly set the distortion level to achieve a required compression rate (as opposed to most other lossy compression techniques mentioned in Section \ref{related work}). A detailed discussion regarding fits' quantization is presented in Section \ref{lossy compression}.  

Notice that while it is customary to consider the leafs as the position of the fits in a tree, in many popular decision tree implementations (such as Matlab's {\sf fitrtree, fitctree,treebagger}, for example), each node of the tree holds a fit, in case of missing values during prediction. This means that the compression rate of the fits takes a significant part in the compressed forest.

\section{Our suggested algorithm}  
\label{Our suggested algorithm}
As described in previous sections, our suggested compression technique decomposes a tree into three components, which are the structure of the tree, the nodes of the tree and the fits of the tree. Since the trees are independent and identically distributed (as a result of the random forest construction) we may compress the trees as memoryless draws from a complex random source, as described in Section \ref{compression methodology}.
Our suggested algorithm works as follows:
We first extract the Zaks sequences which describe the structure of the trees. As mentioned in Section \ref{tree structure compression}, we compress each of these sequences with an LZ-based encoder. We then extract the empirical probability distributions for the nodes names and split values. Specifically, we go over all the nodes in the trees and for each node we record its variable name and split value, its depth in the tree and its father's variable name. We then aggregate this information into a set of conditional empirical probability distributions 
\begin{align}
\nonumber
P_{vn}=P( \text{variable names} |& \text{node depth},\\\nonumber
&\text{father's variable name})
\end{align}
\begin{align}
\nonumber
P_{cv}=P(\text{split value} | \text{node depth, variable name},\\\nonumber
\text{father's variable name}).
\end{align}
Once we have gathered these sets of conditional distributions, we  apply our clustering technique (\ref{clust3}) on ${P}_{vn}$ and ${P}_{cv}$ (separately), to find the ideal tradeoff between a minimal cost of dictionaries' description and minimal averaged redundancy, resulting in using unified dictionaries. We repeat the clustering process for different values of $K$ to find the minimizer of (\ref{clust3}) over all possible $K$'s. Once we have established the chosen clustering and the mean of each cluster (which is a probability distribution $Q_k$), we construct a Huffman code according to $Q_k$ and compress all the cluster's sequences accordingly.  
Lastly, we repeat the same construction of conditional probability distributions to the fits in the tree. We again apply our clustering technique and compress the fits accordingly.  Notice that for two-class classification problems we would usually prefer to use an arithmetic encoder, which tends to out-perform the Huffman encoder for binary alphabets with skewed probability distributions. Algorithm \ref{alg:algorithm} summarizes our suggested method.

\begin{algorithm}[!htbp] 
\caption{Lossless compression of random forests}
\begin{algorithmic} [1]
\REQUIRE A set of $A$ random forest tress, $\{t_1,\ldots,t_A\}$, $v =$ variables names, $d=$number of variables, $C(v_i)=$ set of split values for each $v_i \in v$ and $T =$ maximal depth among all the trees $\{t_1,\ldots,t_A\}$.
\STATE Extract a set of $A$ Zaks' sequences, $\{z_1,\ldots,z_A\}$, from the given trees $\{t_1,\ldots,t_A\}$.
\STATE Concatenate $\{z_1,\ldots,z_A\}$ to a single sequence, $z_{all}$.
\STATE Compress $z_{all}$ using an LZ encoder to achieve $z_{comp}$.

\STATE Set sequences of variable names $vars(dp,fa)=\{ \}$ and corresponding counters $P_{vars}(vn,dp,fa)=0$ for all $dp \in \{1,\dots,T\}$ and $vn,fa \in v$. 
\STATE Set sequences of split values $splits(vn, dp,fa)=\{ \}$ and corresponding counters $P_{spt}(sp, vn,dp,fa)=0$.
\STATE Set sequences of fits $fits(dp,fa)=\{ \}$ and corresponding counters $P_{fits}(vn,dp,fa)=0$.

\FORALL{$t_i\in \{t_1,\ldots,t_A\} $}
\FORALL{$node_j\in t_i $}
\STATE Set $dp=$ the depth of $node_j$'s in $t_i$.
\STATE Set $fa=$ the variable name of $node_j$'s father.
\STATE Set $vn=$ the variable name of $node_j$.
\STATE Set $sp,ft=$ the split and fit values of $node_j$.
\STATE Set $vars(dp,fa)=vars(dp,fa)||vn$.
\STATE Set $P_{vars}(vn,dp,fa)=P_{vars}(vn,dp,fa)+1$.
\STATE Set $splits(vn,dp,fa)=splits(vn,dp,fa)||sp$.
\STATE Set $P_{spt}(sp,vn,dp,fa)=P_{spt}(sp,vn,dp,fa)+1$.
\STATE Set $fits(dp,fa)=fits(dp,fa)||ft$.
\STATE Set $P_{fits}(ft,dp,fa)=P_{fits}(ft,dp,fa)+1$.

\ENDFOR
\ENDFOR

\STATE Normalize all $P$'s by their sums.

\FORALL{$k \in \{1,\dots,K\}$} \label{clust_start}
\STATE Apply the clustering algorithm (\ref{clust3}) with $k$ clusters on the set $P_{vars}$. \label{clust3-alg}
\STATE Set $obj =$ the objective attained in line \ref{clust3-alg} .
\IF {$obj<min{\_}obj$}
\STATE Set $min{\_}obj=obj$, $k{\_}opt = k$. 
\STATE Set $C_{cl}=$ the set of clusters attained in line \ref{clust3-alg}.
\STATE Set $P_{cl}=$ the cluster centers attained in line \ref{clust3-alg}.
\ENDIF
\ENDFOR
 
\STATE set $vars_{comp}=\{\}$.
\FORALL{$k \in \{1,\dots,k{\_}opt\}$}
\STATE Construct a Huffman encoder ${HF}_{vars}(k)$ to $P_{cl}(k)$. 
\FORALL{$P_{vars}\in C_{cl}(k)$}
\STATE Encode the corresponding $vars$ sequence according to ${HF}_{vars}(k)$, to attain $vars{\_}seq_{comp}$.
\STATE Set $vars_{comp}=\{vars_{comp}, vars{\_}seq_{comp}\}$.
\ENDFOR
\ENDFOR \label{clust_end}

\STATE Repeat steps \ref{clust_start} : \ref{clust_end} for $\{P_{splits}\}_{j=1}^d$   to attain the sets of compressed sequences $\{splits_{comp}\}_{j=1}^d$ and corresponding sets of Huffman encoders  $\{{HF}_{splits}\}_{j=1}^d$.

\STATE Repeat steps \ref{clust_start} : \ref{clust_end} for $P_{fits}$ with an arithmetic encoder to attain the set of compressed fits $fits_{comp}$ and a corresponding set of  $P_{fits{\_}cl}$ for decompression purpose.

\RETURN $z_{comp}$, $vars_{comp}$, ${HF}_{vars}$, $\{splits_{comp}\}_{j=1}^d$, $\{{HF}_{splits}\}_{j=1}^d$,  $fits_{comp}$,  $P_{fits{\_}cl}$.
\end{algorithmic}
\label{alg:algorithm}
\end{algorithm}

\section{Predictions from the compressed forest}
As mentioned above, our suggested approach allows making predictions straight from the compressed representation of the forest. This is possible due to the prefix property of the Huffman code. Specifically, given a sequence of symbols that are coded by a Huffman code, we may decode a symbol in the sequence without decoding the entire sequence. This way we may access (and decode) only the required information, to make a prediction for a given future observation. Let us demonstrate our prediction scheme. First, we extract the Zaks' sequence of the first tree. This requires storying $2n+1$ bits in the Random Access Memory (RAM) of the system for a tree of $n$ nodes (see Section \ref{tree structure compression}). Then, for every node that we encounter, we access its compressed variable name and split value in the compressed data, and decode them according to their corresponding Huffman code, as described in Section \ref{Clustering of node models}. Notice that this operation only requires the location of both the compressed information and the corresponding dictionaries in our stored data, which is directly due to the prefix property. Finally, we decode the fit of the leaf, using its corresponding Huffman dictionary, in the same manner as above (Section \ref{fits compression}). We repeat this process for each tree in the forest. Notice that the described scheme may also be used to decode the entire forest, and not just to predict from it.

It is important to emphasize that Huffman code guarantees lossless compression, even if the data is not encoded according to its true underlaying probability distribution \cite{cover2012elements}. This property allows us to reduce the number of Huffman dictionaries that are used in our compression scheme, while still allowing a perfect reconstruction and identical predictions to the original random forest.

\section{Experiments}  

We now demonstrate our suggested compression scheme on a variety of data--driven random forests, generated from publicly available real--world datasets (UCI repository\footnote{\url{http://archive.ics.uci.edu/ml}} and   Kaggle\footnote{\url{http://www.kaggle.com/competitions}}). The random forests are trained using Matlab's {\sf treeBagger} routine with $1000$ trees, while the rest of the parameters are set to their default values. We compare our suggested algorithm with two different lossless compression schemes.  The first, denoted as \textit{standard compression}, begins with applying the {\sf compact(tree)} routine on the trained forest. This creates a compact version of the random forest by eliminating redundant information and duplications of information. Then, the compact version is compressed using {\sf gzip}  \cite{deutsch1996gzip}. These steps attain an immediate lossless compression by currently available off-the-shelf tools. However, notice that the {\sf compact(tree)} routine is not designed solely for prediction purposes and maintains several forest attributes which are unnecessary for our prediction--oriented scheme.  Therefore, we further suggest a \textit{light compression} of a random forest, in which we only keep the information necessary for prediction, as listed in the beginning Section \ref{list}, followed by {\sf gzip} compression. This gives us a more relevant reference for our suggested scheme. It is important to notice that we do make some elementary adjustments to the trees prior to the {\sf gzip} compression, such as replacing the  alphabetical strings along the trees with short numerical values. This further enhances the compression rate of the \textit{light compression} scheme.

It is important to mention that in all of our experiments we use a $64$ bit representation for every numerical fit value we represent. This may be considered as an overly conservative approach for lossless compression. However, for the purpose of this work, we prefer to follow the most orthodox interpretation of losslessness, and show we still achieve high compression rates.

We begin the presentation of our results with a case study, in which we compress a random forest trained over Liberty Mutual Group's Property Inspection Prediction dataset.
In this dataset, the goal is to predict a count of hazards or pre--existing damages using the property's information. This enables Liberty Mutual to more accurately identify high risk homes that require additional examination to confirm their insurability. Liberty dataset consists of $50,999$ observations and $32$ confidential variables, of which $16$ and numerical and $16$ are categorical. We train a random forest according to this dataset, as described above. We then apply the standard compression, to attain a compressed size of $733.7$ MB. We further apply the light compression to the same random forest. This results in $215.6$ MB, of which $122.1$ MB are for the fits. Applying our suggested algorithm achieves a total compression size of $142.7$ MB, where $118$MB describe the fits. We immediately notice that in both of these cases the fits hold a very dominant portion of the forest. This is a result of the numerical nature of the fits, as described in Section \ref{fits compression}. Therefore, let us revert Liberty's regression problem into classification by comparing each observation value to the mean of all observation. This means we would now like to classify those homes for which the number of hazards or pre--existing damages is greater than the mean. We train a random forest for the classification problem and again apply the compression schemes described above. The standard compression results in a total of $723.1$ MB, almost as before. However, the light compression now takes only $96.5$ MB, of which $2.54$ MB are for the trees structure, $10.16$MB for the variable names , $81.3$MB describe the split values and $2.54$MB for the fits. Notice that the fits now take the same portion as the tree structure, since each node holds a single binary fit. 

Applying our suggested compression scheme, we get a total of $12.43$ MB which breaks down to $1.81$MB for the structure, $4.02$MB for the variables names, $4.5$MB for the split values,  $1.58$MB for the fits and the reminder for the dictionaries. These results are summarized in Table \ref{table0}.

\begin{table}[ht]
\caption{Liberty Mutual Classification Problem. Compression Size [MB], For $1000$ Trees RF.}
\centering
\renewcommand{\baselinestretch}{1}\footnotesize
\label{table0} 
\centering
\begin{tabular}{|M{0.8cm}|M{0.7cm}|M{0.7cm}|M{0.8cm}|M{0.5cm}|M{0.5cm}||M{0.6cm}|N}  

\hline

Method
&\begin{tabular}{@{}c@{}} Tree \\ struct.  \end{tabular}
&\begin{tabular}{@{}c@{}} Var. \\ names  \end{tabular}
&\begin{tabular}{@{}c@{}} Split \\ values  \end{tabular}
&\begin{tabular}{@{}c@{}} Fits  \end{tabular}
&\begin{tabular}{@{}c@{}} Dict.  \end{tabular}
&\begin{tabular}{@{}c@{}}Total \end{tabular}&\\[16pt]

\hline

\begin{tabular}{@{}c@{}} light \\ comp.  \end{tabular}
&$2.54$
&$10.16$
&$81.3$
&$2.54$ 
&-- 
&$96.5$&\\[14pt]

\hline

\begin{tabular}{@{}c@{}} Our \\ method  \end{tabular}
&$1.81$
&$4.02$
&$4.5$ 
&$1.58$ 
&$0.52$ 
&$12.43$&\\[14pt]

\hline

\end{tabular}
\end{table}

We notice that by reverting the problem into classification, we achieved a reduction of $124.2$ MB, due to the finite (binary) alphabet of the fits. In total, our suggested scheme achieves a compression rate of $1:40$ compared with the standard compression, and a rate of $1:5.2$ compared with the light compression. 

We further analyze our results and notice that for most variables, the clustering results in three separate models which only depend on the depth of the nodes. This means we usually have a single model for low depth nodes, a single model for middle depth nodes and a single model for deeper nodes. Moreover, we notice that the low depth model (closer to the root) is usually very sparse while the deeper model is almost uniformly distributed. This is not surprising since for a large number of observations, the splits which are closer to the root are expected to have much resemblance over different trees, while deeper splits are much more ``random", due to the greedy construction of the trees. This phenomenon is observed for the variable names models and the split value models.   
Notice that the number of models also strongly depends on the cost of describing each line in the dictionary (the $\alpha$ term in (\ref{clust3})). Since we choose a $64$ bit representation, the cost of a dictionary is relatively large and results in a small number of models. Reducing the representation accuracy to $32$ bits shows an increase in the number of clusters to approximately $7$. 

In addition to Liberty's dataset, we examine our suggested scheme on a variety of classification (marked with $*$) and regression (marked with $+$) problems of different size and complexity. Notice that several classification datasets were generated from regression datasets, as in the Liberty example discussed above. The results are summarized in Table \ref{table1}. All of the datasets are obtained from UCI repository and Kaggle. 

\begin{table}[h]
\caption{Compression Results of $1000$ Trees Random Forests, Trained Over Different Datasets}
\centering
\renewcommand{\baselinestretch}{1}\footnotesize
\label{table1} 
\centering
\begin{tabular}{|M{1.3cm}|M{1.1cm}||M{1cm}|M{1cm}|M{1.1cm}|N}  

\hline

Dataset (method)
&\begin{tabular}{@{}c@{}} \# obs, \\ \# vars  \end{tabular}
&\begin{tabular}{@{}c@{}} standard \\ comp. \\(MB) \end{tabular}
&\begin{tabular}{@{}c@{}} light \\ rep. \\(MB) \end{tabular}
&\begin{tabular}{@{}c@{}} our comp.\\ scheme\\ (MB) \end{tabular}&\\[20pt]

\hline

\begin{tabular}{@{}c@{}} Iris$^*$ \\ ($3$ class) \end{tabular}   
&$150, 4$
&$3.73$
&$0.082$ 
&$\bm{0.013}$&\\[18pt]  

\hline

\begin{tabular}{@{}c@{}} Wages$^*$  \end{tabular}   
&$534, 11$
&$15.78$
&$1.4$ 
&$\bm{0.16}$&\\[18pt]  

\hline

\begin{tabular}{@{}c@{}} Airfoil \\Self Noise$^+$  \end{tabular}   
&$1503, 5$
&$1.364$
&$0.49$ 
&$\bm{0.34}$&\\[18pt]  

\hline

\begin{tabular}{@{}c@{}} Airfoil \\Self Noise$^*$  \end{tabular}   
&$1503, 5$
&$1.26$
&$0.108$ 
&$\bm{0.012}$&\\[18pt]  

\hline

\begin{tabular}{@{}c@{}} Bike\\Sharing$^+$  \end{tabular}   
&$10886, 11$
&$7.69$
&$3.39$ 
&$\bm{2.38}$&\\[18pt]  

\hline

\begin{tabular}{@{}c@{}} Naval\\Plants$^+$  \end{tabular}   
&$11934, 16$
&$8.6$
&$3.05$ 
&$\bm{2.15}$&\\[18pt]  

\hline

\begin{tabular}{@{}c@{}} Naval\\Plants$^*$  \end{tabular}   
&$11934, 16$
&$8.5$
&$2.21$ 
&$\bm{0.81}$&\\[18pt]  

\hline

\begin{tabular}{@{}c@{}} Shuttle$^*$  \end{tabular}   
&$14500, 9$
&$2.162$
&$0.28$ 
&$\bm{0.049}$&\\[18pt]  

\hline

\begin{tabular}{@{}c@{}} Forests$^*$  \end{tabular}   
&$15120, 55$
&$9.136$
&$2.91$ 
&$\bm{0.34}$&\\[18pt]  

\hline

\begin{tabular}{@{}c@{}} Adults$^*$  \end{tabular}   
&$48842, 14$
&$159.1$
&$41.6$ 
&$\bm{7.3}$&\\[18pt]  

\hline

\begin{tabular}{@{}c@{}} Liberty$^+$  \end{tabular}   
&$50999, 32$
&$733.7$
&$215.6$ 
&$\bm{142.7}$&\\[18pt]  
\hline

\begin{tabular}{@{}c@{}} Liberty$^*$  \end{tabular}   
&$50999, 32$
&$723.1$
&$96.5$ 
&$\bm{12.43}$&\\[18pt]  

\hline

\begin{tabular}{@{}c@{}} Otto$^*$  \end{tabular}   
&$61878, 94$
&$209.1$
&$48.3$ 
&$\bm{6.1}$&\\[18pt]  
\hline

\end{tabular}
\end{table}

As we can see, our suggested scheme achieves an average compression rate of approximately $1:70$ compared with the standard compression and approximately $1:6$, compared with the light compression, for the classification problems. However, the average compression rates for the regression problem are only $1:4.1$ and $1:1.45$ compared with the two compression methods respectively, as a result of the costly lossless compression of the numerical fits, as discussed above.  In most of the datasets, the model clustering results in $2-3$ different models, in the same manner as in the Liberty dataset. This further justifies the relaxation of our trees' model, as described in Section \ref{Our suggested algorithm}, so that in practice there is no need for exponentially growing number of models prior to the clustering phase.  

\section{Lossy compression}
\label{lossy compression}
Although the focus of our work is lossless compression of random forests, there are several immediate adjustments which allow a lossy compression with favorable theoretical guarantees.
In this section we introduce two basic lossy modifications, which are tree sampling and fits quantization.

Let $A$ be a set of independent and identically distributed trees, trained by the random forest routine, over a dataset of $n$ observations. Let $A_0$ be a randomly sampled subset of $A$. We would like quantify the accuracy loss and the compression gain, caused by the sampling operation. 

Notice that while it is customary to regard the observations as random entities (for generalization purposes), in the context of data compression we regard them as fixed. Therefore, the randomness of the ensemble is solely due to the forest construction routine.

For each observation $i$, denote the mean random forest prediction for this observation on this specific dataset by $\hat{y}^*_i$. Denote the prediction from a random tree $t\in A$ in the random forest sequence by $\hat{y}_{t,i}$, and the ``error" it incurs by $e_t(i) = \hat{y}_{t,i}-\hat{y}^*_i$.
Let $\mu_i$ and $\sigma^2_i$ be the mean and variance of this error, respectively. Let us now randomly sample a subset $A_0 \subset A$ of the ensemble. Then, the accuracy loss may be bounded from above by 
$$ D(A,A_0,\sigma^2)=\text{var} \left(\frac{\sum_{t \in A_0} e_t}{|A_0|} - \frac{\sum_{t \in A} e_t}{|A|}   \right)$$
where $e_t$ is the mean of $e_t(i)$ for all $t \in A$. Notice that the random variables $e_t$ are i.i.d. with a mean $\mu=n^{-1}\sum_{i=1}^n \mu_i$ and a variance $\sigma^2_i$. We assume for simplicity that  $\sigma^2_i =\sigma^2$ is fixed (or that $\sigma^2_i$ is bounded from above by $\sigma^2$). Then $\text{var}(e_t)$ is between $\frac{\sigma^2}{n}$ and $\sigma^2$, depending on the dependence structure between predictions of the same tree. However, since $\text{var}(e_t)<\sigma^2$, we have that $\sigma^2>\sigma^2_i \quad \forall i \in \{1\dots n\}$. Simple derivation shows that
\begin{align}
D(A,&A_0,\sigma^2)=\\\nonumber
&\text{var} \left(\frac{\left(|A| |A_0|\right)\sum_{k \in A_0} e_k}{|A| |A_0|} -\frac{\sum_{k \in A \ A_0}}{|A|}    \right)=\\\nonumber
&\sigma^2 |A_0| \left(\frac{1}{|A_0|}+\frac{1}{|A|}\right)^2+\sigma^2\frac{|A|-|A_0|}{|A|^2}.
\end{align}
Assuming that $|A_0|\ll|A|$ we have that $$D(A,A_0,\sigma^2) \approx \frac{\sigma^2}{|A_0|}+\frac{\sigma^2}{|A|}.$$
It is important to mention that even though our derivation considers the ensemble's trees, $t \in A$,  as a random entity, in practice they are regarded as fixed data structures to be compressed. This means that the $\sfrac{\sigma^2}{|A|}$ term is the ``ground truth" of our random forest prediction accuracy and the accuracy loss, caused by sampling $|A_0|$ trees (followed by lossless compressing), is simply $\sfrac{\sigma^2}{|A_0|}$. Assuming that subsamping the ensamble does not effect the compression rate of individual trees, the compression gain we achieve is fairly straight forward and shown to be linear in the sampling ratio, $\nicefrac{|A_0|}{|A|}$, on the average.

On top of subsampling the trees, an additional lossy compression adjustment may be attained through quantizing the (numerical) fits, as discussed in Section \ref{fits compression}. Assume the fits take values over a finite range of size $2^c$. Let us quantize the values of the fits with a naive $b$ bits quantization. This means we define $2^b$ quantization points and uniformly place them over the range. Assuming that the distortion (quantization error) is uniformly distributed (for example, through dithered quantization \cite{schuchman1964dither}) we attain an average accuracy loss of $\sfrac{2^r}{2^b}=2^{-(b-r)}$. Further, assuming that each numerical value is represented by $64$ bits, the compression gain we achieve is $\sfrac{b}{64}$, on the average.

Therefore, the average overall accuracy loss (that is, the variance of the difference before and after subsampling $|A_0|\ll|A|$ trees and quantizing the numerical fits)  is bounded from above by
$$ \frac{\sigma^2}{|A_0|}+\frac{\left(2^{-(b-r)}\right)^2}{12|A_0|},$$
while the average compression gain is a factor of $\sfrac{b}{64}$ for the compressed fits and an additional factor of $ \sfrac{|A_0|}{|A|}$ for the entire compressed ensemble. Notice that while there exist more adequate frequency based quantization techniques (for example, Lloyd-max \cite{lloyd1982least}), the naive quantization described above offers simple and favorable theoretical properties. However, in practice, one may achieve better performance by applying those methods. 

\begin{figure}[!ht]
\centering
\includegraphics[width = 0.52\textwidth,bb= 120 180 500 610,clip]{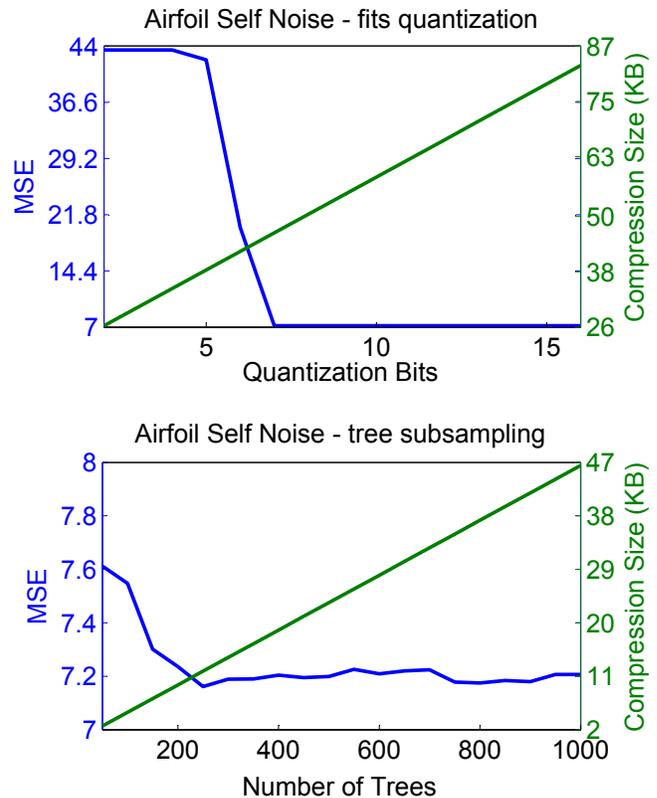}
\caption{Lossy compression of Air Foil Noise data-set. Upper chart: fits quantization. Lower chart: tree subsampling. The Blue curve is the MSE on the test-set and the green curve is the corresponding compression size }
\label{lossy_fig_1} 
\end{figure}

Let us now illustrate our suggested lossy compression approach. Figure \ref{lossy_fig_1} demonstrates the fits quantization (upper chart) and the tree subsampling (lower chart), applied to the (regression) Air Self Noise data-set (see Table \ref{table1}). Here, we split the data-set to $80\%$ train-set and $20\%$ test-set. We train a random forest (using Matlab's {\sf treebagger} routine) and evaluate the mean square error (MSE) on the test-set.  Then, we apply the two lossy compression techniques discussed above. The upper chart demonstrates the effect of the fits quantization. The $x$-axis is the number of quantization bits used to describe the fits, the blue curve is the corresponding MSE (on the test-set) and the green curve is the compression size. As we can see, we may represent the fits by only $7$ bits, with no significant degradation in performance of the random forest. This results in a compression size of approximately $47$ KB. Notice that the over-conservative $64$ bit representation used in Table \ref{table1} allows a compression size of $340$ KB.  Now, let us subsample the trees in the forest, while maintaining the $7$ bits representation for the fits. The lower chart of Figure  \ref{lossy_fig_1} demonstrates the MSE (blue curve) and the resulting compression size (green curve) for different numbers of subsampled trees ($x$-axis).  Here we observe that by sampling only $250$ trees of the forest, we may reduce the compression size to only $11$ KB, while almost maintaining the same performance. Therefore, we conclude that by both quantizing the fits and subsampling the forest, we may reduce the compression size from $340$ KB (in the conservative lossless case) to only $11$ KB with no significant impact on the generalization performance. In addition, we notice the linear threads of our compression size curves, which illustrate (and justify) our analysis above.

\begin{figure}[!ht]
\centering
\includegraphics[width = 0.52\textwidth,bb= 110 180 520 610,clip]{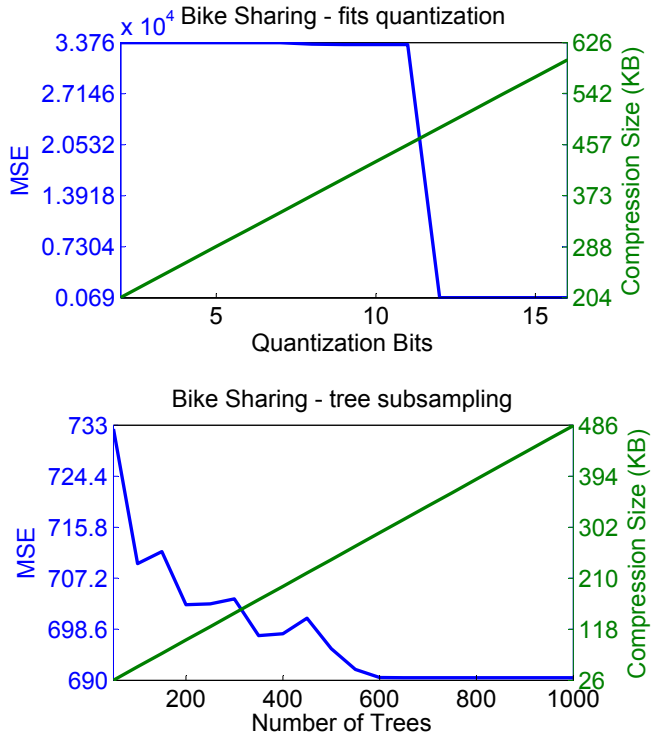}
\caption{Lossy compression of the Bike Sharing data-set. Upper chart: fits quantization. Lower chart: tree subsampling. The Blue curve is the MSE on the test-set and the green curve is the corresponding compression size}
\label{lossy_fig_2} 
\end{figure}

Let us further apply our lossy compression techniques to a larger data-set. Figure \ref{lossy_fig_2} demonstrates the MSE and the corresponding compression size of our suggest method, applied to the (regression) Bike Sharing data-set (Table \ref{table1}). Here, we may reduce the compression size from $2.38$ MB to only  $300$KB with no significant effect on the generalization performance. This is achieved by representing the fits with $12$ bits, while subsamping $600$ trees from the forest.

It is important to mention that our suggested lossy approach is typically not competitive with some alternative methods, such as neural-networks based compression \cite{bucilua2006model}. Our suggested lossy compression typically compresses the forest in a factor of up to a $100$ (from the uncompressed representation), while neural-based methods compress in factors of $1000$ and more \cite{bucilua2006model}. However,  our main advantage lies in the ability to provide a theoretically sound trade-off between distortion and compression rate and to explicitly control the desired performance. In addition, our method allows to further modify the forest (for example, by adding more trees), even after the lossy compression is applied.  This serves as a balancing mechanism for coding implementations.

\section{Discussion and conclusion}
In this work we introduce a novel method for lossless compression of random forests. Our suggested method uses the independent and identically distributed nature of the trees to fit probabilistic models and compress the data accordingly. Since the number and the complexity of the models grow with the size of the problem, we apply model clustering according to Bregman divergence. This allows us to find the optimal trade-off between a smaller set of models that accurately describe the data, and corresponding dictionaries for decompression purposes. 

While to the best of our knowledge, our suggest approach is unique in its lossless nature, there exists a large body of work on lossy compression of ensemble methods. Most of these lossy compression schemes manipulate the forest (by pruning or mimicking it), with hardly any guarantees on the resulting prediction accuracy. The main advantage of our suggested scheme is that it provides a complete and accurate recovery of the forest. This property ensures the same prediction accuracy as the original forest. In addition, it allows future modification to the forest (such as adding more trees, applying further inference, etc.). Further, since our method is lossless and directly compresses the trees, a more complex random forest would not necessarily result in a worse compression rate (as demonstrated in Table \ref{table1}). Notice that the lossy schemes, on the other hand, may result in a severe deterioration of accuracy in order to achieve a prescribed compression rate, as described in Section \ref{related work}.

Although the focus of our work is lossless compression, our suggested scheme may also extend to lossy compression, as described in Section \ref{lossy compression}. The main advantage of our lossy scheme is that it is easy to implement and provides theoretical guarantees on both the accuracy loss and the achieved compression gain. This allows the user to find the ideal balance between the two without blindly applying a series of lossy compression tasks.

It is important to mention several popular variants of tree ensembles which imply different probabilistic structures. For example, Completely Randomized Trees (CRT) \cite{geurts2006extremely,liu2008spectrum} utilize a recursive partitioning in which the observations in each node are split according to a randomly chosen feature and a corresponding random split value. Therefore, we expect less resemblance among the trees. Further, it leads to more uniform distributions of the splitting rules in each node, and henceforth, a lower compression rate. On the other hand, there exist more complicated tree-based structures such as Deep Forest \cite{zhou2017deep}, where different random forests are cascaded layer-by-layer, similarly to Deep Neural Networks. This results in more involved probabilistic dependencies, as we consider the collection of all the trees in the system. Nevertheless, we may still cluster and encode different models together, to introduce a compression gain. 

All of these properties make our suggested compression framework a favorable methodology, both in theory and in practice.

\section*{Acknowledgments}
This research was partially supported by Israel Science Foundation grant 1487/12 and by a returning scientist fellowship from the Israeli Ministry of Immigration to Amichai Painsky.

\bibliographystyle{IEEEtran}
\bibliography{bibi}

\end{document}